\documentclass{article}

\usepackage{microtype}
\usepackage{graphicx}
\usepackage{subcaption}
\usepackage{booktabs} 

\usepackage{hyperref}
\usepackage{xurl}
\usepackage[utf8]{inputenc}
\usepackage[T1]{fontenc}

\usepackage{needspace}

\usepackage{fancyvrb}  
\usepackage{fvextra} 
\usepackage[most]{tcolorbox} 

\usepackage[accepted]{icml2026}

\usepackage{amsmath}
\usepackage{amssymb}
\usepackage{mathtools}
\usepackage{amsthm}

\usepackage[capitalize,noabbrev]{cleveref}

\theoremstyle{plain}

\theoremstyle{definition}

\theoremstyle{remark}

\usepackage[textsize=tiny]{todonotes}

\icmltitlerunning{Reinforcement Learning for Tool-Calling Agents in Fast Healthcare Interoperability Resources (FHIR)}

\begin{document}

\twocolumn[
  \icmltitle{\textbf{Reinforcement Learning for Tool-Calling Agents in Fast Healthcare Interoperability Resources (FHIR)} }

  \icmlsetsymbol{equal}{*}

  \begin{icmlauthorlist}
    \icmlauthor{Marius S Knorr}{equal,yyy}
    \icmlauthor{Robert Müller}{equal,comp}
    \icmlauthor{Jan P Bremer}{yyy}
    \icmlauthor{Nils Schweingruber}{yyy}
  \end{icmlauthorlist}

  \icmlaffiliation{yyy}{IDM gGmbH, University Medical Center Hamburg-Eppendorf, Hamburg, Germany}
  \icmlaffiliation{comp}{Aganthos}

  \icmlcorrespondingauthor{Marius Knorr}{knorr@idmedizin.de}

  \icmlkeywords{Machine Learning, ICML}

  \vskip 0.3in
]

\printAffiliationsAndNotice{\icmlEqualContribution}

\begin{abstract}

Fast Healthcare Interoperability Resources (FHIR) is the dominant standard for interoperable exchange of healthcare data. In FHIR, electronic health records form a directed graph of resources. Answering clinically meaningful questions over FHIR requires agents to perform multi-step reasoning, filtering, and aggregation across multiple resource types. Prior work shows that even tool-augmented LLM agents (retrieval, code execution, multi-turn planning) often select the wrong resources or violate traversal constraints. We study this problem in the context of FHIR-AgentBench, a benchmark for realistic question answering over real-world hospital data, and frame reasoning on FHIR as a sequential decision-making problem over a queryable structured graph. We implement a multi-turn CodeAct agent and post-train it with reinforcement learning using a custom harness and tools. A LLM Judge provides execution-grounded rewards. Compared to prompt-based, closed-model baselines, RL post-training improves performance while enforcing data-integrity constraints. Empirically, our approach improves answer correctness from 50\% (o4-mini) to 77\% on FHIR-AgentBench using a smaller and cheaper Qwen3-8B model. We present an end-to-end post-training pipeline (environment building, harness construction, model training and custom evaluation) that reliably improves multi-turn reasoning over structured clinical graphs.

\end{abstract}
\begin{figure*}[t]
  \centering
  \includegraphics[width=\linewidth]{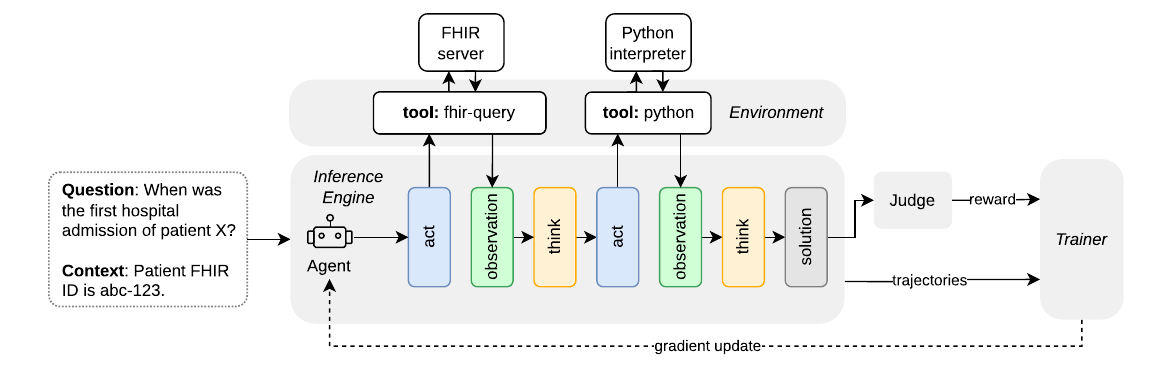}
  \caption{
  Reinforcement learning pipeline for training a clinical FHIR reasoning agent. The agent follows a structured act–observe–think loop, with access to two tools: a retrieval tool for loading clinical resources from a FHIR server, and a Python interpreter for code execution. After one or more reasoning cycles, the agent emits a final solution, which is evaluated by a LLM-judge against ground-truth answers from FHIR-AgentBench to produce a reward signal. The rewards and full interaction trajectories are then fed to the trainer, which updates the agent's policy via GRPO, closing the RL loop.
  }
  \label{fig:flowchart-figure}
\end{figure*}

\section{Introduction}
\label{sec:introduction}

The broad availability of electronic health data \citep{adler-milstein_electronic_2017, kim_challenges_2024} holds potential to enable a wide range of data-driven applications by integrating signals across different data modalities. A primary barrier to realizing this potential is not only the high dimensionality and incompleteness of EHR (electronic health records) data, but its lack of standardization across institutions, vendors, and workflows \citep{reisman_ehrs_2017}. As a result, leveraging clinical data at scale frequently requires site-specific schema alignment through one-off ETL pipelines, limiting reproducibility and portability.

To address interoperability, the health IT ecosystem has increasingly converged on Fast Healthcare Interoperability Resources (HL7 FHIR) \citep{hl7_fhir_r4_4_0_1}, a data model for clinical and administrative data with a RESTful web API for exchange and access. In FHIR, entities are represented as resources (e.g., \textit{Patient}, \textit{Encounter}, \textit{Observation}, \textit{MedicationRequest}) that may reference one another. Consequently, a patient chart can be viewed as a directed graph with nodes and edges. 

Clinically meaningful information is rarely captured in a single resource but instead distributed across several linked nodes. For example, the fact that a patient was prescribed a particular medication and later showed elevated creatinine spans separate \textit{MedicationRequest} and \textit{Observation} resources connected through shared references to an \textit{Encounter} and \textit{Patient}. Making sense of such data therefore requires multi-hop traversal across these resources, together with operations such as filtering or temporal sorting. This turns clinical question answering into a graph navigation problem. 

\citet{lee2025fhiragentbenchbenchmarkingllmagents} target this challenge with \textsc{FHIR-AgentBench}, a benchmark of 2,931 clinician-sourced questions grounded in real de-identified patient records from MIMIC-IV converted to FHIR format.
The benchmark evaluates whether LLM agents can navigate the FHIR graph, retrieve the relevant data, and reason over it to produce correct answers. Their systematic evaluation compares retrieval strategies (direct FHIR API calls vs.\ specialized tools), interaction patterns (single-turn vs.\ multi-turn), and reasoning strategies (natural language vs.\ code generation). The results expose two core bottlenecks: on the retrieval side, agents select wrong resource types or miss required traversal steps; on the reasoning side, they mishandle temporal logic, case-sensitive terminologies, or multi-step aggregation over the retrieved data. Even in their strongest configuration (o4-mini with multi-turn planning and code execution), accuracy reaches only approximately 50\%, showing the gap between current agent capabilities and the demands of structured clinical reasoning over FHIR.

We hypothesize that this brittleness stems from a gap between FHIR’s standardized interfaces and real-world deployments. In practice, optional fields are populated inconsistently, and the same clinical concept can be encoded in different resources or profiles. Prior work has shown that the use of \textit{Observation} is particularly inconsistent across implementation guides, though this heterogeneity extends beyond \textit{Observation} \citep{kramer_reducing_2023}. Single-shot LLMs that assume an idealized schema therefore struggle when faced with real-world FHIR data. In contrast, a multi-turn interaction pattern can support schema discovery: the agent can inspect candidate resources, verify how resources are stored, and iteratively refine queries before committing to an answer \citep{lee2025fhiragentbenchbenchmarkingllmagents}. 

To operationalize this idea, we adopt a CodeAct-style agent paradigm \citep{wang2024executablecodeactionselicit} in which actions are executable programs that query the FHIR API, inspect intermediate results, and self-correct across turns. We implement this multi-turn code agent within SkyRL, a post-training harness that supports both evaluation of open-weight models and specialization for FHIR QA through execution-based rewards (Figure~\ref{fig:flowchart-figure}).

Building on \textsc{FHIR-AgentBench} \citep{lee2025fhiragentbenchbenchmarkingllmagents}, we show that post-training with multi-turn tool use enables the agent to internalize the specific structure and conventions of the underlying FHIR server through repeated interaction and debugging.

Empirically, we improve accuracy from 50\% (o4-mini) to 77\% using a substantially smaller Qwen3-8B model. We further provide a practical post-training recipe and an analysis of failure modes, highlighting the remaining gaps toward reliable FHIR-based clinical QA.

\section{Background}
\label{sec:background}

\subsection{FHIR}
\label{sec:fhir_overview}

\begin{figure*}[t]
  \centering
  \includegraphics[width=\linewidth]{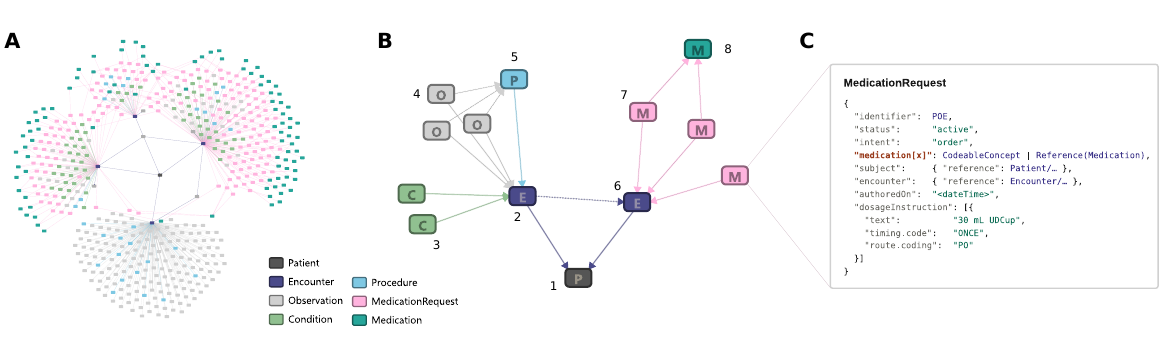}
  \caption{
    Fast Healthcare Interoperability Resources (FHIR).  \textbf{(A)} Resource graph of a MIMIC-FHIR patient, connected by references. \textbf{(B)} Clinical concepts can be mapped to FHIR: A patient (1) presents to the emergency department (2; first Encounter) with one or more conditions (3). During the visit, clinical observations (4) were recorded (e.g. heart rate, blood pressure), each linked to the procedure that generated them (5, e.g. vital signs measurement). The patient was subsequently admitted to the hospital (6; second Encounter). During the inpatient stay, Furosemide (h) was requested at a dose of 40 mg (g). Notice that one of the three MedicationRequest resources does not reference a Medication resource. Instead, the medication information is contained in the MedicationResource itself (contained resource). 
    }
  \label{fig:fhir-intro-plot}
\end{figure*}

Fast Healthcare Interoperability Resources (FHIR) is an HL7 standard for representing clinical and administrative data as a collection of typed resources (e.g., \textit{Patient}, \textit{Encounter}, \textit{Observation}, \textit{MedicationRequest}) exchanged via a RESTful API. Each resource is a self-contained JSON document identified by its type and a unique ID, and contains nested, domain-specific attributes. Resources reference one another through typed Reference fields. For instance, an \textit{Observation} points to the \textit{Patient} it describes, forming a typed, directed, heterogeneous graph (Figure~\ref{fig:fhir-intro-plot}) that can be traversed and queried incrementally.

From an ML viewpoint, a patient record is therefore not a single
table but a graph of heterogeneous nodes with sparse, inconsistently
populated attributes; answering a clinical question requires
programmatic traversal and computation over this structure.

FHIR standardizes resource types and interaction patterns but intentionally
leaves room for local specialization. Deployments routinely constrain base
resources through \emph{profiles} (\textit{StructureDefinition}), introduce
site-specific fields via \emph{extensions}, and adopt different terminology
systems for the same clinical concepts. This flexibility is essential for
adoption, but it creates a core ML challenge: two sites can represent the same clinical fact using different resource types, fields, or coding conventions, while optional elements may be populated inconsistently.

\subsection{Related work}
\label{sec:related_work}

We find it useful to separate prior work into two pipelines: \emph{(i) populating} a FHIR server (mapping raw clinical data into FHIR resources), and \emph{(ii) reading} a FHIR server (retrieving/traversing resources to answer questions or complete tasks).

\emph{(i)} A line of work uses LLMs to convert clinical narratives or structured datasets into FHIR-compliant resources. \citet{li_fhir-gpt_2024} propose FHIR-GPT, targeting medication-focused extraction by transforming free-text snippets into \textit{MedicationStatement} resources and evaluating against an annotated dataset. More recently, \citet{riquelme2025largelanguagemodelsautomating} study LLM-assisted transformation of structured clinical tables into FHIR on MIMIC-IV, using retrieval and schema-aware prompting to map tabular fields to resource attributes. Complementary to these model-centric efforts, \citet{idrissi-yaghir_using_2025} introduce FHIR-Workbench, a suite of datasets that includes note-to-FHIR generation and resource recognition tasks, providing standardized evaluation for both write and read capabilities.

\emph{(ii)} A different line of work treats a FHIR store as the backend for question answering or interactive clinical agents. \citet{schmiedmayer2024llmfhirdemystifying} built an application that lets users query FHIR-formatted patient data with an LLM through a mobile application (synthetic data). \citet{kothari2025questionansweringpatientmedical} follow a retrieve-then-read pipeline: they first retrieve relevant FHIR resources for a query and then answer using a fine-tuned (private) LLM, focusing on patient-record QA with privacy constraints (using synthetic data via Synthea). Benchmarks have recently matured from static QA sets to interactive, tool-using evaluations. \citet{jiang2025medagentbenchrealisticvirtualehr} introduce \textit{MedAgentBench}, a FHIR-compliant virtual EHR environment with multi-step, clinician-authored tasks spanning diverse categories (beyond QA). Our work builds directly on FHIR-AgentBench \citep{lee2025fhiragentbenchbenchmarkingllmagents}, which grounds thousands of clinician-sourced questions in real MIMIC-IV-FHIR records and systematically evaluates single-turn vs. multi-turn agents, natural-language vs. code-based reasoning, and different retrieval interfaces. We adopt their benchmark as our starting point.

\subsection{Tool-Calling LLM Agents}
\label{sec:tool_calling_agents}

Tool-calling LLM agents are commonly evaluated not only on whether they select the correct tool, but also on whether they emit executable calls and recover from downstream errors. Recent benchmarks emphasize multi-turn and stateful function-calling settings \citep{patil2025bfcl}, while stable benchmarking efforts address the practical instability of real-world APIs and evaluation randomness \citep{guo2024stabletoolbench}. A consistent finding is that tool use is fragile: small deviations in argument values or formatting can trigger tool failures that propagate through a toolchain, even when the high-level intent is correct \citep{xiong2025butterfly}. This fragility is amplified in multilingual interactions, where models may generate semantically correct parameter values in the user's language that violate execution conventions \citep{luo2026lostexecution}. Importantly, the \emph{semantic specification} of tools (names, signatures, and natural-language descriptions) shapes the effective action space and reduces ambiguity in tool selection and argument construction, making it a key determinant of reliable tool use \citep{müller2025semanticcontexttoolorchestration}.

These failure modes are directly relevant to EHR backends such as FHIR, where strict schemas and traversal constraints make tool-call validity a first-order concern. A single malformed search parameter or an incorrectly typed reference field can silently return empty results, derailing an otherwise correct reasoning chain.

\subsection{Reinforcement Learning for Tool Use}
\label{sec:rl_tool_use}

Privacy and data-security constraints in clinical settings often motivate on-premise deployment with open-weight models, which typically underperform closed models on realistic EHR tool-use benchmarks without specialization \citep{lee2025fhiragentbenchbenchmarkingllmagents}. Reinforcement learning from verifiable feedback/rewards (RLVF/RLVR) offers a practical route to close this gap. Group Relative Policy Optimization (GRPO), introduced in DeepSeekMath and adopted in DeepSeek-R1 \citep{shao2024deepseekmathpushinglimitsmathematical,deepseek-ai_deepseek-r1_2025}, demonstrated that strong reasoning can emerge from automatically checkable signals, initially in mathematical domains \citep{deepseek-ai_deepseek-r1_2025}. Follow-up analyses further study when and why such RLVR-style training succeeds \citep{liu_understanding_2025}. The same principle extends to tool use, where execution outcomes provide a natural verifiable reward: recent work applies GRPO-style post-training to strategic tool integration, execution-grounded optimization, and tool-centric agentic reasoning \citep{feng2025retoolreinforcementlearningstrategic,singh2025agenticreasoningtoolintegration,zhang2025nemotronresearchtooln1exploringtoolusinglanguage,qian2025toolrlrewardtoollearning}.

More recent work extends this line along several axes.
On the \emph{training} side, TAPO optimizes adaptive
tool-calling policies that interleave reasoning with
on-demand invocation of search and code tools
\citep{wu2025tapo}; Fission-GRPO converts execution
errors into on-policy corrective supervision rather than
treating them as sparse negative rewards
\citep{zhang2026fissiongrpo}; and ToolOrchestra trains
compact 8B-parameter orchestrators to allocate calls
across heterogeneous models and tools under outcome-,
efficiency-, and preference-aware objectives
\citep{su2025toolorchestra}.
Orthogonally, reliability can be improved without
additional training by constraining tool interaction
through Hoare-style contracts that gate invocation on
verified preconditions and commit results only after
runtime postcondition checks \citep{liu2026toolgate}.
On the \emph{data} side, complementary pipelines
bootstrap tool learning by synthesizing verifiable hard
cases from agent failure graphs \citep{hao2026hardgen},
converting exploratory trajectories into tasks that
stress-test ambiguous and evolving user intents
\citep{wang2026trajectory2task}, and deriving virtual
training trajectories from real tool calls with
mutation-based negative samples to reduce intent
deviation \citep{xiong2026rise}.

In structured EHR backends such as FHIR, answers can be verified against the ground-truth record, making the setting a natural fit for RLVR-style post-training.

\subsection{Interpreter-Mediated Reasoning}
A line of work frames agent actions as executable programs rather than atomic tool calls, delegating precise computation to an external interpreter. PAL~\cite{gao2023pal} and Program-of-Thoughts~\cite{chen2022pot} generate code to separate natural-language reasoning from arithmetic, while ReAct~\cite{yao2023react} popularizes the general interleaving of reasoning and acting with environment feedback. CodeAct~\cite{wang2024executablecodeactionselicit} instantiates this pattern with Python code as action. SkyRL-SQL~\cite{liu2025skyrlsql} demonstrates that such agents benefit from RL post-training with execution-grounded rewards in multi-turn Text-to-SQL; our setting is analogous but structurally different, applying RL post-training to CodeAct agents over schema-constrained FHIR graphs.

\section{Method}
\label{sec:method}

\subsection{Problem definition}
\label{sec:problem}

A patient's record in FHIR forms a typed, directed, heterogeneous graph
(\S\ref{sec:fhir_overview}). Answering a clinical question over a FHIR graph requires selecting resource types, traversing references across hops, and applying client-side computation
(filtering, temporal alignment, aggregation, etc.).
We focus on the \emph{reading} side of interoperability: given access to an
existing FHIR server, how can an agent reliably retrieve and reason over the
information needed for a correct answer?

The central difficulty is that FHIR standardizes interfaces but not content:
local profiles, extensions, and terminology choices mean the same clinical
concept can appear in different resources or fields, and optional elements are
populated inconsistently across sites~\citep{kramer_reducing_2023}.
A single-shot query plan that assumes a fixed schema is therefore brittle.
We instead adopt an iterative pattern, \emph{schema discovery}, in which
the agent probes candidate resources, inspects returned JSON, and refines its
strategy before committing to an answer.

\subsection{System architecture}
\label{sec:architecture}

We implement post-training with
\textsc{SkyRL}~\citep{griggs2025skrylv01}, which provides an end-to-end RL
pipeline comprising (i)~a trainer (we use GRPO-style updates;
\S\ref{sec:training}), (ii)~a high-throughput rollout engine based on
\textsc{vLLM}, and (iii)~an environment that defines tool interfaces, parses
actions, and returns observations.

Episodes follow a ReAct-style loop: after each observation (tool output or
error message), the agent produces a brief reasoning step and then either
issues a tool call or outputs a final answer.
We cap each episode at $N_{\max}=6$ turns, where a turn is either a single tool
invocation or a final answer.

\paragraph{Environment and tools.}
The environment exposes two tools and a terminal action:
\begin{enumerate}
  \item A \emph{Retrieval} tool takes a patient ID and a resource type and queries all resources of that type for the given patient.
     For resource types not directly linked to \textit{Patient} (e.g.,
    \textit{Medication}), the tool first fetches patient-linked
    \textit{MedicationRequest} resources and then resolves the referenced
    \textit{Medication} entries.
    The results are stored in a shared workspace, accessible for the Python tool.

  \item A \emph{Python} tool executes Python code over the retrieved resources, enabling both schema inspection (by printing samples into context) and arbitrary data manipulation (filtering, sorting, temporal alignment, unit conversions, aggregation, etc.).
  \item Finally, a \emph{Finish} action carries the agent's final answer and terminates the episode. It is exposed as a tool for interface uniformity but does not interact with the environment.
\end{enumerate}

The retrieval tool is inspired by \citet{lee2025fhiragentbenchbenchmarkingllmagents}; we intentionally prioritize simplicity over efficiency and leave optimized query planning to future work.
Actions are emitted as XML-tagged tool calls in the native Qwen3 format and executed by the environment; for non-terminal actions, the resulting output is fed back as the next observation.

\paragraph{FHIR server.}
We deploy a \textsc{Blaze} FHIR server~\citep{samplyBlaze2025} and populate it
with the $100$ MIMIC-FHIR demo patients used by \textsc{FHIR-AgentBench}.
If not stated otherwise, results are reported on the 424 \textsc{FHIR-AgentBench} validation set questions.

\paragraph{Prompting.}
The system prompt was intentionally kept
short:
\begin{quote}\small\ttfamily
You are a FHIR data analyst. Answer patient data questions by querying a
FHIR server. Rules:
\begin{itemize}
  \item Every claim must trace to a \textnormal{\texttt{print()}} output or computation.
  \item If unsure about a resource's schema, print a sample first.
  \item Keep your reasoning brief.
  \item When done, call finish.
\end{itemize}
\end{quote}
The prompt is followed by tool descriptions (Appendix~\ref{sec:system-prompt}), which are serialized into the system prompt by the Qwen3 chat template.

\subsection{Training}
\label{sec:training}

\paragraph{Formulation.}
Each episode begins with a clinical question $q$ and an execution context $c$
(patient identifier, time horizon, task-specific constraints).
At step $t$ the agent observes the question and the full interaction history
$h_t = (q,\, c,\, a_1, o_1, t_1,\dots, a_{t-1}, o_{t-1}, t_{t-1})$
and selects an action $a_t$, i.e.\ a code snippet executed by the runtime
(\S\ref{sec:architecture}). The runtime returns observation $o_t$ (retrieved resource counts, printed information, or error
messages), which is appended to the history. Each $t_i$ denotes an optional thinking block produced by the agent after observation $o_i$. An initial thinking step $t_0$ may optionally precede the first action.
This multi-turn structure is what enables schema discovery: early actions probe
candidate resources and inspect how information is recorded, while later
actions exploit those findings to retrieve and aggregate the answer. 

A trajectory has the form
$\tau = (q,\, c,\, a_1, o_1, t_1,\dots, a_T, o_T, t_T, y)$, where $y$ is the final
natural-language answer emitted at step~$T$.
An episode terminates when the agent produces $y$ or exceeds the turn budget
$N_{\max}$ or token budget $L_{\max}$.
We optimize expected answer correctness:
\begin{equation}\label{eq:objective}
  \max_\theta\;
  \mathbb{E}_{(q,c) \sim \mathcal{D},\;
              \tau \sim \pi_\theta(\cdot \mid q,c)}
  \bigl[\, r(\tau) \,\bigr],
\end{equation}
where $\pi_\theta$ is the agent policy (an LLM parameterized by~$\theta$) and
$r(\tau)\!\in\!\{0,1\}$ indicates whether $y$ is correct.

\paragraph{LLM judge.}
We use Qwen2.5-72B-Instruct as an automatic judge, following the
evaluation protocol and prompting scheme of \citet{lee2025fhiragentbenchbenchmarkingllmagents}.
The trajectory receives $r(\tau)=1$ if the output matches the required answer
format \textbf{and} the predicted answer is judged correct; otherwise
$r(\tau)=0$.

\paragraph{GRPO.}
We optimize Eq.~\ref{eq:objective} with Group Relative Policy Optimization
(GRPO;~\citealp{shao2024deepseekmathpushinglimitsmathematical}), which
eliminates the need for a learned value function by estimating advantages from
a group of sampled trajectories.
For each query $q$ the current policy $\pi_{\theta_{\mathrm{old}}}$ samples a
group of $G$ trajectories $\{\tau_i\}_{i=1}^G$ with corresponding binary
rewards $\{r_i\}_{i=1}^G$.
The advantage of the $i$-th trajectory is:
\begin{equation}\label{eq:advantage}
  \hat{A}_i
  = \frac{r_i - \mathrm{mean}(\{r_j\}_{j=1}^G)}
         {\mathrm{std}(\{r_j\}_{j=1}^G)}.
\end{equation}
GRPO maximizes a clipped surrogate with a KL penalty toward a reference policy
$\pi_{\mathrm{ref}}$ which we update every epoch with the latest policy weights:
\begin{equation}
\begin{split}
J_{\mathrm{GRPO}}(\theta) & = \mathbb{E}_{q \sim \mathcal{D},\;
    \{\tau_i\} \sim \pi_{\theta_\mathrm{old}}(\cdot\mid q)}
   \Bigg[\,
   \frac{1}{G}\sum_{i=1}^{G}
\\
&\quad
   \frac{1}{\lvert \tau_i \rvert}
   \sum_{t=1}^{\lvert \tau_i \rvert}
   \Bigl(
     \min\bigl(
       \rho_{i,t}(\theta)\,\hat{A}_i,
\\
&\quad
     \mathrm{clip}\bigl(\rho_{i,t}(\theta),\,
                        1{-}\epsilon_{\mathrm{low}},\,
                        1{+}\epsilon_{\mathrm{high}}\bigr)\,
     \hat{A}_i
     \bigr)
\\
&\quad
   - \beta\,D_{\mathrm{KL}}\!\bigl(\pi_\theta \,\|\, \pi_{\mathrm{ref}}\bigr)
   \Bigr)
   \,\Bigg],
\end{split}
\end{equation}
with importance ratio
$\rho_{i,t}(\theta)
 = \pi_\theta(\tau_{i,t} \mid q,\,\tau_{i,<t})\;/\;
   \pi_{\theta_{\mathrm{old}}}(\tau_{i,t} \mid q,\,\tau_{i,<t})$,
where $\tau_{i,t}$ denotes the $t$-th agent-generated token in
trajectory~$\tau_i$.

Each completion $\tau_i$ in our setting spans multiple interaction turns:
it comprises all agent-generated tokens (code snippets and the final answer)
across the episode; environment observations are excluded from the likelihood.
The reward $r_i$ is assigned once at episode end, so the per-token advantage
$\hat{A}_i$ is constant across all tokens in trajectory~$i$.
During training, multiple queries form a batch and gradients are averaged.

\section{Experiments}
We use instruction-tuned Qwen3 models with native tool calling and reasoning.

\subsection{Baselines}
FHIR-AgentBench reports an answer correctness of 50\% with o4-mini in a multi-turn CodeAct setting \citep{lee2025fhiragentbenchbenchmarkingllmagents}. In our harness, o4-mini achieves a comparable 47\%. We additionally include Gemini-3-Flash (52\%) and GLM-5 (59\%, 744B parameters with 40B active) as API-based baselines. To isolate the contribution of RL training (see \cref{sec:rl-experiments-8b}), we evaluate Qwen3 across four model sizes (4B, 8B, 14B, 32B) in a zero-shot setting. For each prompt, we perform five inference passes per prompt (5×424 rollouts total; temperature 0.1, 12-turn budget), reporting the mean score, standard deviation, and pass@5 in \cref{tab:baselines}.

\begin{figure}[ht]
    \includegraphics[width=\columnwidth]{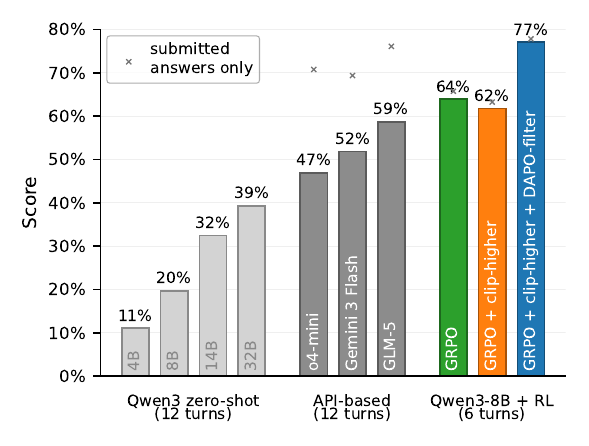}
    \caption{Answer correctness on FHIR-AgentBench \citep{lee2025fhiragentbenchbenchmarkingllmagents} of vanilla Qwen3 models in different sizes (4B to 32B), API-based models (closed/open-weights), and after RL training (step 3000). Small crosses indicate performance restricted to questions where the agent successfully submitted an answer.}
    \label{fig:baselineplot}
\end{figure}

\subsection{RL Training}
\label{sec:rl-experiments-8b}

Qwen3-8B was trained with a fixed learning rate of $1 \times 10^{-6}$ and a batch size of~8. For variance reduction, we use a group size of~8 (i.e., eight rollouts per prompt) and restrict to a 12{,}000-token context length. Train rollouts were sampled with a temperature of~1.0, whereas a temperature of~0.1 was used for evaluation rollouts. We ran three RL recipes:

\begin{enumerate}
    \item GRPO baseline with symmetric clipping ($\epsilon_{\mathrm{low}} = \epsilon_{\mathrm{high}}=0.2$) and token-mean normalization.
    \item Like 1., but with asymmetric clipping (``clip-higher''; $\epsilon_{\mathrm{low}}=0.2$, $\epsilon_{\mathrm{high}}=0.28$) instead of symmetric clip.
    \item Like 2., but with dynamic sampling as proposed by the DAPO paper \citep{yu_dapo_2025}: zero-variance groups (all correct or all incorrect) are filtered out from training.
\end{enumerate}

\begin{figure}[ht]
    \includegraphics[width=\columnwidth]{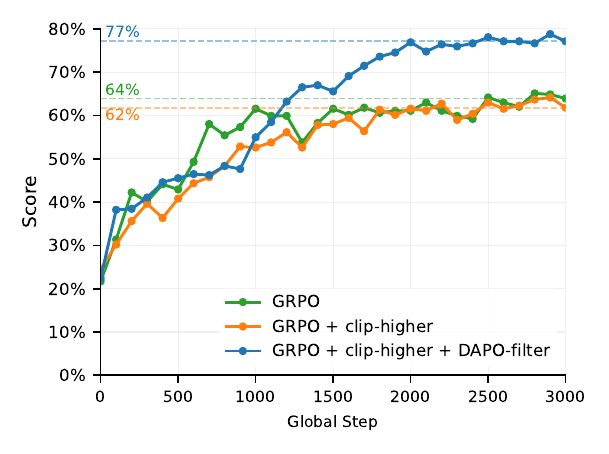}
    \caption{Training curves for Qwen3-8B. Curves show the answer correctness on the FHIR-AgentBench \citep{lee2025fhiragentbenchbenchmarkingllmagents} validation subset (n=424) with a 6 turn budget. Recipes: vanilla GRPO (blue), +clip-higher (orange), +DAPO-filter (green). Dashed horizontal lines mark each run's score at step 3000.
    }
    \label{fig:learning-curve}
\end{figure}

\begin{figure}[ht]
    \includegraphics[width=\columnwidth]{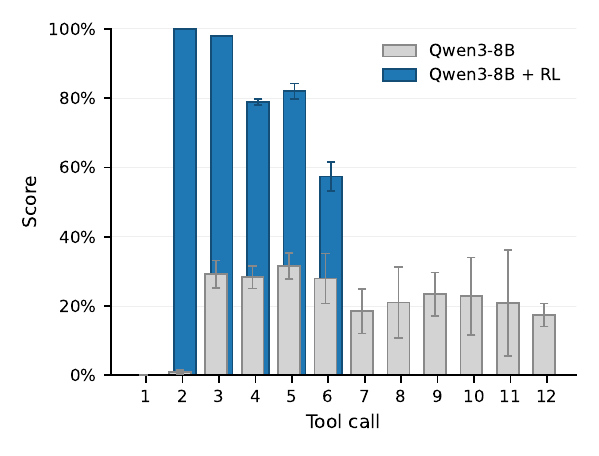}
    \caption{Per-turn score for vanilla Qwen3-8B (12-turn budget, grey) vs. GRPO-trained Qwen3-8B at step 3000 (6-turn budget, blue). Error bars (±1 SD) capture variability across rollouts within each bin, with five rollouts per prompt.}
    \label{fig:fewer-turns-plot}
\end{figure}

Symmetric and asymmetric clipping show nearly identical performance after 3000 training steps. Answer correctness reaches 64\% for setup~1 and 62\% for setup~2 (Figure~\ref{fig:learning-curve}). Given this negligible difference, we introduce dynamic sampling as a third recipe. Setup~3 reaches 77\% correctness in fewer steps, suggesting that filtering zero-variance groups improves sample efficiency. We further observe that RL training reduces the number of turns needed to submit a solution, i.e., the model learns to answer within a six-turn budget (Figure~\ref{fig:fewer-turns-plot}). This shows a clear effect of RL training beyond mere elicitation: answer correctness increases while the number of turns required decreases.

\begin{table}[t]
  \caption{Answer correctness on FHIR-AgentBench (\%). Mean $\pm$ SD computed over 5 rollouts for Qwen3 models.}
  \label{tab:baselines}
  \begin{center}
    \begin{small}
      \begin{sc}
        \begin{tabular}{lcc}
          \toprule
          Model & Score & pass@5\\
          \midrule
          Qwen3-4B           & 11.1 $\pm$ 1.3 & 29.5 \\
          Qwen3-8B           & 19.7 $\pm$ 0.3 & 47.2 \\
          Qwen3-14B          & 32.5 $\pm$ 1.0 & 53.8 \\
          Qwen3-32B          & 39.3 $\pm$ 1.3 & 68.6 \\
          \midrule
          o4-mini            & \multicolumn{1}{l}{46.9} & --   \\
          Gemini 3 Flash     & \multicolumn{1}{l}{51.9} & --   \\
          GLM-5              & \multicolumn{1}{l}{58.7} & --   \\
          \midrule
          Qwen3-8B + RL$^\dagger$ & 76.7 $\pm$ 0.9 & 79.7 \\
          \bottomrule
        \end{tabular}
      \end{sc}
    \end{small}
    \vskip 0.05in
    {\footnotesize $^\dagger$GRPO + clip-higher + DAPO-filter}
  \end{center}
  \vskip -0.1in
\end{table}

\subsection{Judge}
\label{sec:judge}

We measured agreement between the automated judge (Qwen2.5-72B-Instruct) and human annotations from a board-certified physician. Annotations cover the 424 validation samples from the base-GRPO run (setup 1) at step 3000. The resulting agreement is $96.2\%$ (precision $= 96.7\%$, recall $= 97.4\%$), in line with the 97\% reported by \citet{lee2025fhiragentbenchbenchmarkingllmagents}.

For the API-based baselines, we use the same Qwen2.5-72B-Instruct model served via the OpenRouter API in fp8 precision. Agreement between the OpenRouter judge and our local vLLM judge (fp16) was measured at $99.75\%$, confirming that the quantization difference has negligible impact.



\subsection{Breakdown}
\begin{figure}[ht]
\includegraphics[width=\columnwidth]{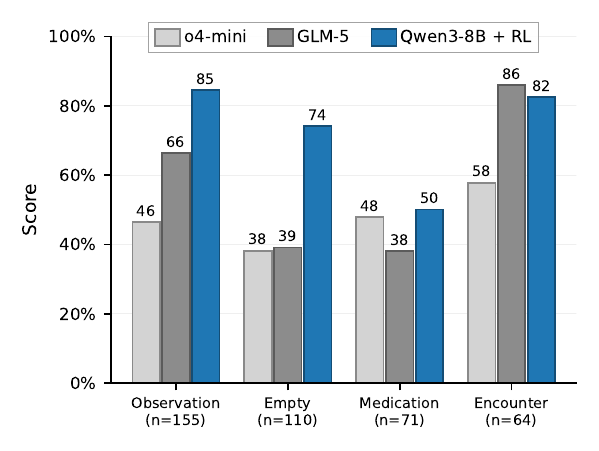}
\caption{Answer correctness by FHIR resource type. Empty refers to questions with no matching FHIR resources (i.e., negative or null answers). Categories with fewer than 15 validation samples (Location, Procedure, Patient, Condition) are omitted.}
\label{fig:taskbreakdown}
\end{figure}

To analyze failure modes, we categorize validation questions by the ground-truth FHIR resource type required to answer them, which is annotated in the dataset (Figure~\ref{fig:taskbreakdown}). Among the 424 questions in the validation split, most require fetching Encounter ($n=64$), Observation ($n=155$), Medication/MedicationRequest ($n=71$), or belong to the Empty category ($n=110$). 

All questions require filtering resources of the relevant type by patient as a first step, using the query tool. For Encounter questions, the agent performs Python post-processing such as selecting relevant fields or sorting by date. Observation questions additionally require string-matching the \textit{code.coding[0].display} field to locate a specific lab value (or other values), followed by filtering analogous to the Encounter case. Medication questions require resolving the MedicationRequest-Medication reference: the agent must fetch both resource types and surface the Medication ID from MedicationRequest into context in order to dereference it (see Figure~\ref{fig:trajectoryfig}). The Empty category contains questions whose ground-truth FHIR ID list is empty, meaning the agent must conclude that no matching record exists; reward is given only if the agent reports a negative or zero answer. 

The agent performs worse on questions that require following the Medication reference. While the agent succeeds on some prompts (Figure~\ref{fig:trajectoryfig}), it does not apply this skill reliably. We hypothesize that longer training would improve performance further: the skill has been acquired but is applied inconsistently.

\begin{figure}[ht]
  \includegraphics[width=\columnwidth]{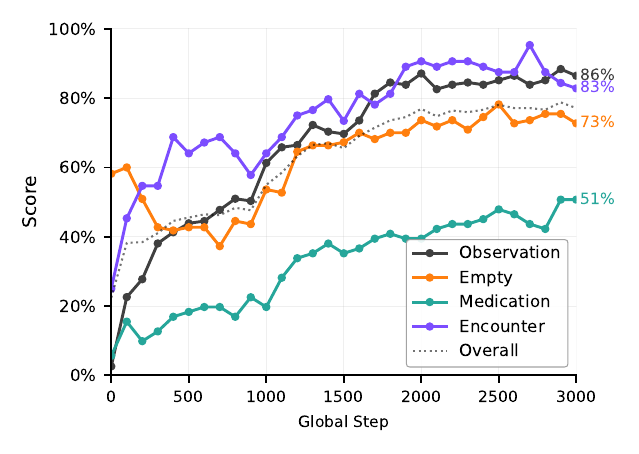}
  \caption{Our best RL run (GRPO + clip-higher + DAPO filter), broken down by FHIR resource type over training steps. The model first learns to return "Empty" on questions with no matching resources, which is the cheapest reward to get, before improving on the other resource types. The dotted line shows the overall mean.}
  \label{fig:taskbreakdown_trainingcurve}
\end{figure}

For the best RL run (setup 3: GRPO + clip-higher + DAPO-filter), we also show the resource-type breakdown over training steps (Figure~\ref{fig:taskbreakdown_trainingcurve}). The agent first learns to commit to a no-answer, which is the easiest reward signal to get. Eventually, it learns to answer questions that require interacting with the data, with some negative transfer to the Empty category. Overall performance improves with training, though spikes in one curve often coincide with drops in others.

\section{Conclusions}
\label{sec:conclusions}
Privacy constraints often make on-premise deployment a necessity for clinical AI; however, smaller open-weight models typically perform worse on realistic EHR tasks than their closed-source counterparts. We show that with the right post-training recipe, a small open model can exceed closed-model performance on structured clinical reasoning.

We frame FHIR question answering as a sequential decision-making problem over a typed, heterogeneous resource graph, and instantiate it as a multi-turn CodeAct agent that probes schema variants, retrieves resources, and processes JSON in Python. The agent is post-trained with GRPO using execution-grounded rewards from an LLM-judge, validated at 96\% agreement with a board-certified physician.

The results suggest that the limiting factor for open models on FHIR was not capacity but specialization: the trained agent learns the conventions of the underlying server through repeated interaction and debugging, rather than relying on an idealized schema. 

Inspection of trajectories indicates that most remaining failures involve reference traversal across resource types, e.g., resolving MedicationRequest-Medication references, which the agent does not yet do consistently. Future FHIR benchmarks could target this capability directly by including harder questions that require traversing three or more resource types.

Several limitations qualify this picture. Our retrieval interface fetches whole resource types and will not scale efficiently to large patient populations without a planning layer for targeted FHIR search. Evaluation depends on an LLM judge. While agreement with our physician annotator is high (96\%), the remaining disagreement bounds how precisely further improvements can be measured. Finally, while 77\% answer correctness is a substantial gain, it remains well below the reliability bar required for high-stakes clinical use.

Future work should test whether the recipe transfers to larger backbones (e.g., 32B) and replace the simple retrieval tool with learned query planning to scale to larger populations. 

More broadly, our findings indicate that execution-grounded RL post-training is a viable path to capable on-premise clinical agents, closing the open-versus-closed gap in exactly the setting where on-premise deployment is most needed.

\section*{Impact statement}
This work contributes to improving the robustness of large language model agents for structured clinical data reasoning, with potential to support more scalable and interoperable healthcare analytics. In clinical settings, agent errors carry concrete consequences: a missed medication or misattributed lab value could mislead downstream decisions. Our best system reaches 77\% answer correctness, well below the bar for autonomous clinical use, and we view this work as a step toward decision-support tools that require human verification, not replacement of clinical judgment.

\bibliography{references}
\bibliographystyle{icml2026}


\newpage
\appendix
\onecolumn
\section{Appendix}




\definecolor{promptbg}{HTML}{F5F5F5}
\definecolor{promptborder}{HTML}{C8C8C8}
\definecolor{prompttitle}{HTML}{FFFFFF}

\newtcolorbox{promptbox}[1][]{
  enhanced,
  colback=promptbg,
  colframe=promptborder,
  boxrule=0.5pt,
  arc=2pt,
  left=6pt, right=6pt, top=4pt, bottom=4pt,
  fonttitle=\bfseries\footnotesize\sffamily,
  coltitle=black,
  attach boxed title to top left={xshift=8pt, yshift=-6pt},
  boxed title style={
    colback=prompttitle,
    colframe=promptborder,
    boxrule=0.5pt,
    arc=1pt,
    left=4pt, right=4pt, top=1pt, bottom=1pt,
  },
  title=#1,
  top=12pt,
}

\subsection{System prompt}
\label{sec:system-prompt}
The system prompts for all experiments consist of two parts. First, the natural-language instruction:
\begin{promptbox}[System prompt pt. I]
\begin{Verbatim}[fontsize=\footnotesize, breaklines=true,
                 breakanywhere=true, breaksymbolleft={}, breaksymbolright={}]
You are a FHIR data analyst. Answer patient data questions by querying a FHIR server.
Rules:
- Every claim must trace to a print() output or computation.
- If unsure about a resource's schema, print a sample first.
- Keep your reasoning brief.
- When done, call finish.
\end{Verbatim}
\end{promptbox}

Second, the tool schemas. Since we use native tool calling, every chat template serializes the tool schema differently. The Qwen3 chat template serializes the tool schemas into the following string:

\newcommand{\toolname}[1]{\textbf{#1}}

\begin{promptbox}[System prompt pt. II]
\begin{Verbatim}[fontsize=\footnotesize, breaklines=true,
                 breakanywhere=true, breaksymbolleft={}, breaksymbolright={},
                 commandchars=\!\@\?]
# Tools
You may call one or more functions to assist with the user query.
You are provided with function signatures within <tools></tools> XML tags:
<tools>
{"type": "function",
 "function": {
   "name": "!toolname@fhir_query?",
   "description": "Query a FHIR server for health records. Retrieved resources are accumulated across calls. Use multiple calls to gather all the data you need before answering.",
   "parameters": {
     "type": "object",
     "properties": {
       "resource_type": {"type": "string",
         "description": "FHIR resource type to query, e.g. Patient, Condition, Observation, MedicationRequest, Procedure, etc."},
       "patient_fhir_id": {"type": "string",
         "description": "Patient FHIR ID to filter by."}},
     "required": ["resource_type", "patient_fhir_id"]}}
}
{"type": "function",
 "function": {
   "name": "!toolname@python?",
   "description": "Execute Python to analyze or transform FHIR data.\n`retrieved_resources` (dict of resource_type -> list[dict]) only contains\ndata from prior fhir_query calls - fetch before you analyze.\nOnly printed output is returned.",
   "parameters": {
     "type": "object",
     "properties": {
       "code": {"type": "string",
         "description": "Python code to execute. Use print() to produce output."}},
     "required": ["code"]}}
}
{"type": "function",
 "function": {
   "name": "!toolname@finish?",
   "description": "Signals the completion of the current task or conversation.\n\nUse this tool when:\n- You have successfully completed the requested task\n- You cannot proceed further due to technical limitations or missing information\n\nThe answer field should include the final answer to the problem (follow the required format) if an answer is required by the problem.\n",
   "parameters": {
     "type": "object",
     "properties": {
       "answer": {"type": "string",
         "description": "Final message summarizing the task or containing the answer."}},
     "required": ["answer"]}}
}
</tools>
For each function call, return a json object with function name and arguments within <tool_call></tool_call> XML tags:
<tool_call>
{"name": <function-name>, "arguments": <args-json-object>}
</tool_call>
\end{Verbatim}
\end{promptbox}

\begin{figure*}[t]
  \centering
  \includegraphics[width=0.95\linewidth]{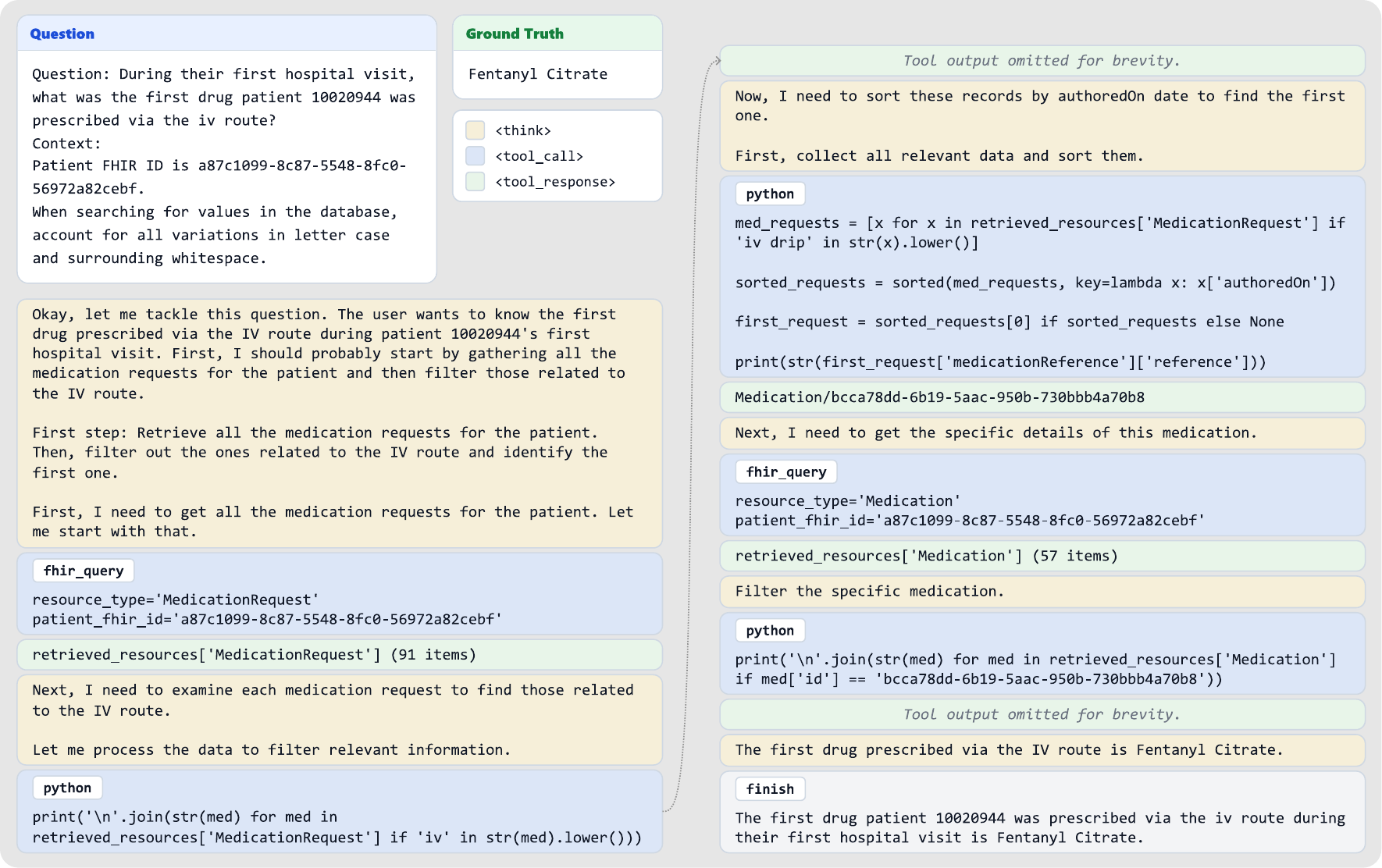}
  \caption{
  Example trajectory. The agent first retrieves all MedicationRequest resources. Then it filters for "iv", and sorts the resources. It identifies the correct MedicationRequest resource, and then resolves the reference to Medication. Medication holds the actual medication that the question is looking for. In this plot, Python tool outputs were collapsed for brevity.
  }
  \label{fig:trajectoryfig}
\end{figure*}

\begin{figure*}[t]
  \centering
  \includegraphics[width=0.95\linewidth]{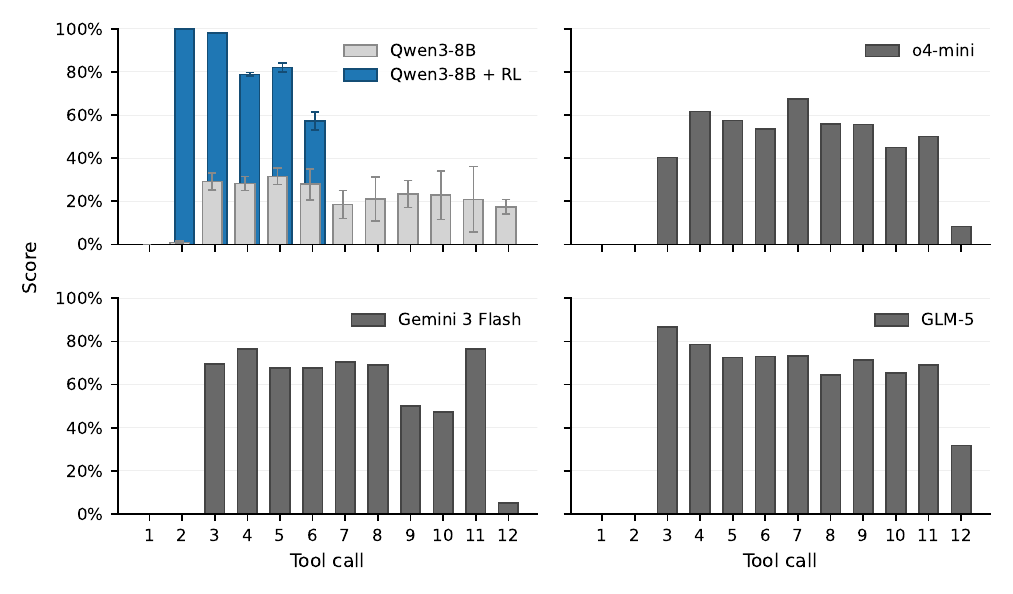}
  \caption{Accuracy by FHIR resource type for Qwen3 (zero-shot and trained), and API-baselines.}
  \label{fig:scorebytoolcallAlsoAPImodels}
\end{figure*}


\end{document}